\documentclass{article}
\usepackage[utf8]{inputenc}
\usepackage{authblk}
 \usepackage[numbers]{natbib} %for Vancouver style in bibliography
\usepackage{graphicx}
\usepackage{amsmath}
\usepackage{comment} %%to comment long streaks of text
\usepackage{caption}
\usepackage{lscape} %%for horizontal tables
\usepackage{subcaption}
\usepackage{multirow}
\usepackage{hyperref} %%for url
\usepackage{setspace}   %Allows double spacing with the \doublespacing command
\usepackage{tabularx,ragged2e,booktabs,caption} %%for table title
\newcolumntype{C}[1]{>{\Centering}m{#1}}  %%table title
\title{LEXpander: applying colexification networks to automated lexicon expansion}
\author[1,2,3]{Anna Di Natale\footnote{Corresponding author. Email: dinatale@csh.ac.at, Telephone: +43 159991604}}
\author[2,1,3]{David Garcia}
\affil[1]{Center for Medical Statistics, Informatics and Intelligent Systems, Medical University of Vienna. Spitalgasse 23, BT88, 1090 Vienna, Austria.}
\affil[2]{Faculty of Computer Science and Biomedical Engineering, Graz University of Technology. Inffeldgasse 16c/I, 8010 Graz, Austria.}
\affil[3]{Complexity Science Hub Vienna. Josefstädter Straße 39, 1080 Wien, Austria}

\date{}

\begin{document}
\maketitle

\doublespacing

\newpage
\section{Abstract}
%Current NLP applications to social media text and historical corpora rely on word lists to detect topics or to measure subjective states.
\subsection*{Background / Introduction}
Recent approaches to text analysis from social media and other corpora rely on word lists to detect topics, measure meaning, or to select relevant documents. These lists are often generated by applying computational lexicon expansion methods to small, manually-curated sets of root words. Despite the wide use of this approach, we still lack an exhaustive comparative analysis of the performance of lexicon expansion methods and how they can be improved with additional linguistic data.
\subsection*{Methods}
In this work, we present LEXpander, a method for lexicon expansion that leverages novel data on colexification, i.e. semantic networks connecting words based on shared concepts and translations to other languages. We evaluate LEXpander in a benchmark including widely used methods for lexicon expansion based on various word embedding models and synonym networks.
\subsection*{Results}
We find that LEXpander outperforms existing approaches in terms of both precision and the trade-off between precision and recall of generated word lists in a variety of tests. Our benchmark includes several linguistic categories and sentiment variables in English and German. We also show that the expanded word lists constitute a high-performing text analysis method in application cases to various corpora.
\subsection*{Conclusion}
This way, LEXpander poses a systematic automated solution to expand short lists of words into exhaustive and accurate word lists that can closely approximate word lists generated by experts in psychology and linguistics.
\newline
\\

\textbf{Keywords:} colexification networks, lexicon expansion, text analysis, word embeddings

\section{Introduction}
Lists of words are widely used in many text analysis and NLP tasks, either in a pre-processing step to retrieve relevant instances in texts or as a necessary part of text analysis algorithms, for example in methods relying on word counts. Even the application of methods not based on word lists, for example neural networks, the selection of the texts to retrieve and analyse is often based on some thematic word lists to query in larger corpora. For example, in order to to classify suicide related tweets with machine learning methods, Metzler et al. \cite{metzler2021detecting} first retrieve relevant tweets on the basis of a word list. Furthermore, established benchmarks in text analysis, such as the SemEval benchmark for sentiment analysis \cite{semeval} and the Tweeteval benchmark \cite{tweeteval}, are based on querying large text sources (e.g. the whole of Twitter) by using pre-specified word lists.

Word lists\footnote{We focus on word lists or lexica that are related to a chosen topic or behavior but do not have additional variables or metadata such as valence or sentiment ratings.} dealing with a subject are either created from scratch or adapted from already published word lists used in previous studies. Already existing word lists might be adapted to research questions which are slightly different from the original ones. In many cases the novel setting or the research questions of a new study require a modification of the original word list. These modifications can range from the exclusion of words not suitable for the topic analysed, as in \cite{metzler2021collective, jaidka2020estimating}, to the translation of those word lists into other languages, as in \cite{werlen2021emotions}. However, some of these manipulations could result in the introduction of noise or error in the new version of the word list. Indeed, modification like the translation of expressions and words in a different language might not convey the same meaning as the original ones \cite{mohammad2020practical}. 
Additionally, novel topics have to be addressed with word lists created ad hoc, as for example coronavirus-related word lists at the beginning of the covid-19 pandemic, as in \cite{banda2021large}, or word lists used to explore modes of drug administration not previously known to the medical personnel \cite{balsamo2021patterns}. Extended word lists can be manually created starting from a short selection of seed words and expanding them to create a final word list via brainstorming or with the use of a thesaurus. This approach often proves to be resource-intensive and prone to inconsistent results when replicated by different groups of people \cite{king2017computer}. 

An alternative to manual lexicon expansion is the application of automated methods to find new words. One of the first approaches is the retrieval of synonyms and related words using semantic resources like WordNet \cite{miller1995wordnet}. A more recent approach consists in the use of word embedding spaces to select the closest words to each keyword, as in \cite{balsamo2021patterns}. Indeed, according to the distributional hypothesis the retrieved words are semantically related to the chosen seed words. One of the most elaborated lexicon expansion methods is Empath \cite{fast2016empath}, which deploys word embeddings to generate word lists from a list of keywords. In particular, Empath constructs the expanded word list considering the closest words to the cumulative vector of the seed words in the word embedding space. Although Empath is widely used in many studies \cite{ribeiro2018characterizing,shing2018expert,zirikly2019clpsych}, the method deploys an outdated word embedding model, as the algorithm has not been updated since its release. 

Multiple solutions have been provided to solve the problem of lexicon expansion but researchers lack a systematic comparison of existing methods that allow both to find the best-performing ones and to validate their performance in relevant application scenarios. A notable resource in this extent is Lexifield \cite{mpouli2020lexifield}, a lexicon expansion algorithm that was compared to previous existing methods. While informative, the work on Lexifield is based on a narrow set of methods (word embedding-based and knowledge-based approaches) and on a limited set of topics (sound, taste and odour). Another effort towards the creation of a baseline is constituted by \cite{bozarth2022keyword}. There, the authors investigate the problem in relation to the retrieval of tweets. In this article, we aim to provide a wider benchmark that includes methods based on synonym networks, word embeddings, and colexification networks, to word list cases of wide use such as emotion words, sentiment words, words for behaviors such as cognition and social interaction, as well as words for several topics including family members, religion, death, work and leisure. 
Beyond a benchmark for lexicon expansion, we also present a novel approach to the problem of automatic lexicon expansion: LEXpander. LEXpander is based on a multilingual semantic network of colexification. Colexification is a linguistic phenomenon that occurs when two different concepts are conveyed using the same word in one language. This language is said to colexify the two concepts. For example, the two concepts 'medicine' and 'poison' are expressed with only one word, 'pharmacon', in Ancient Greek. Therefore Ancient Greek colexifies the concepts of 'poison' and 'medicine'. Since its coinage, the concept of colexification has been related to semantic similarity \cite{franccois2008semantic}, that is concepts that are linked by colexification share semantic meaning \cite{xu2020conceptual, karjus2021conceptual}.

Colexification occurrences have been collected from multiple linguistic resources and organized in a network structure, where concepts that are colexified by a number of languages are linked \cite{list2014clics, croft2022two}. The most known colexification network is Clics$^3$ \cite{rzymski2020database} and is built form a wide range of linguistic resources. However, this network presents a small set of concepts (less than 2,000 in the newest version), unsuitable for the application to lexicon expansion. In order to increase the language coverage of Clics$^3$, colexification networks automatically built from bilingual dictionaries have been proposed in our previous work \cite{di2021colexification}. We showed that these automatically built networks encode affective relationships and reach high performance when inferring the affective ratings of words, with the FreeDict colexification network being one of the most comprehensive. 

In this paper, we adapt the inference method based on the FreeDict colexification network from \cite{di2021colexification} to the expansion of lexica, a method we call LEXpander. We use FreeDict instead of other colexification networks because FreeDict has been shown to yield to the best performance when recovering the affective meaning of words \cite{di2021colexification} and because it encompasses the highest number of words, nearly 28,000. In contrast to \cite{di2021colexification}, in this work we do not only focus on the affective dimension of meaning, but aim at developing a tool of lexicon expansion which can be applied to the most various themes. Furthermore, we do not tackle the the problem of the inference of ratings of words. On the contrary, we aim at expanding thematic word lists, that is at selecting words related to a specific theme without specifying the intensity of this relationship.

A novel feature of LEXpander is that, by design, it provides a multilingual solution to the problem of lexicon expansion. The underlying structure of colexification networks is language-independent, which makes the system applicable to any language included in the translation data used to build the colexification network. In this paper, we showcase this feature by testing LEXpander for the automatic expansion of lexica both in English and German. In the literature there are few attempts at expanding word lists in languages different from English. In particular, \cite{mpouli2020lexifield} expands word lists in English and French, \cite{zeng2018chinese} deploys Chinese linguistic resources and \cite{thavareesan2020tamil} consider the expansion of sentiment-related word lists in Tamil. However, German is a language which has not yet been addressed in previous literature of lexicon expansion algorithms.

The contributions of this paper are the following:
\begin{itemize}
    \item We propose a novel lexicon expansion method, LEXpander, which is based on a linguistic concept;
    \item We compare LEXpander with other widely used lexicon expansion algorithms establishing a benchmark for lexicon expansion algorithms;
    \item We show that LEXpander achieves the best precision and $F_1$ when expanding word lists in English and in German;
    \item We show that LEXpander has either the best or is tied with the best method in terms of correlation with exhaustive manual word lists in the analysis of texts of online and traditional communication;
    \item We present an interactive web app and a R package to allow the easy use and extension of LEXpander.
\end{itemize}

\section{Methods}
\subsection{Lexicon expansion algorithms}
\begin{figure}[hbtp]
  \includegraphics[width=\linewidth]{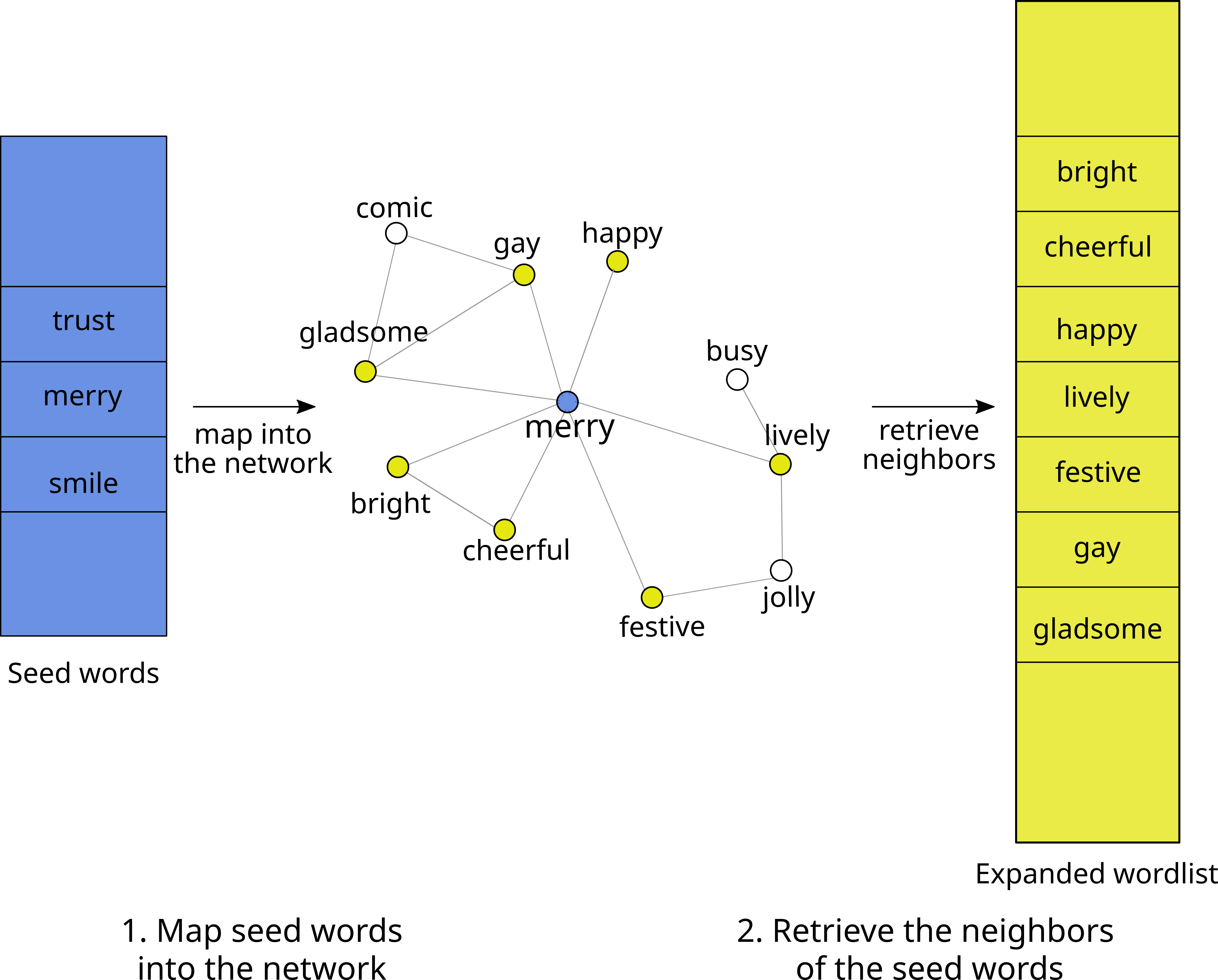}
  \caption{Representation of the LEXpander algorithm. Seed words (blue, on the left) are mapped onto the network (step 1) and the neighbors of those words (yellow, on the right) are retrieved to create the expanded word list (step 2). In the figure, we only represent the process in the case of one word, 'merry'.}
  \label{fig:method}
  \end{figure}

In this paper, we propose a novel method for lexicon expansion, LEXpander, and compare its performance with other approaches. Text analysis applications often rely on the ad hoc creation of lexica which ideally collect all the words used to refer to a topic. LEXpander automatizes the task of creating such word lists starting from a small set of seed words. LEXpander is based on a colexification network, that is a multilingual semantic network whose structure is language independent.\\
The LEXpander model is built as follows. Given the adjacency matrix of the colexification network $A={A_{ij}}$, such that:
\begin{equation*}
    A_{ij}=\begin{cases}
    1 & \text{if concepts $i$ and $j$ are colexified by at least two languages}.\\
    0 & \text{otherwise}
    \end{cases}
\end{equation*}
and $S=\{s\}$ set of seed words, the expanded lexicon is defined as $L=S \cup W$, where:
\begin{equation}
    W=\{w | A_{sw}=1, \forall s \in S\}
    \label{eq:method}
\end{equation}
In other words, given a set of seed words $S$, LEXpander creates a longer lexicon by retrieving all the neighbors of the seed words in the colexification network, as represented in Figure \ref{fig:method}. In order to expand word lists in German, we deploy equation (\ref{eq:method}) with the German version of the colexification network. The German version of the network is obtained translating the labels of the colexification network in German using the FreeDict dictionary (\url{freedict.org}). Note that the translation of the labels does not change the structure of the colexification network, which is thus language-independent from among the languages included in the dataset.\\

We compare the performance of LEXpander with various other automatic lexicon expansion algorithms. These methods can be divided in two classes: methods which deploy word embeddings and methods based on semantic networks. As methods based on semantic networks, we consider the widely-used network of synsets retrieved from WordNet \cite{miller1995wordnet} for the expansion of English word lists and an open source version of WordNet in German, OdeNet \cite{siegel2021odenet}. The lexicon expansion approach which deploys these semantic networks is similar to the one used for LEXpander in (\ref{eq:method}): the expansion of a set of seed words consists in the retrieval of the neighbors of all the seed words in the network. One necessary step of the expansion of lexica using semantic networks is the mapping of the seed words onto the network (see Figure \ref{fig:method}, step 1). However, in some cases the seed words cannot be mapped onto the network and the expansion of the word list is not possible. In this case, the result of the expansion is an empty word list. Moreover, together with the performance of each method we also consider the number of word lists the method could expand as a way to consider this case\footnote{This can also happen in word embedding models, thus we report this for all methods in our benchmark.}.

A second class of methods we consider are approaches based on word embeddings. In particular, we consider methods based on the GloVe model trained on the English Wikipedia \cite{pennington2014glove} and on the German Wikipedia and the FastText model trained on the English Wikipedia and on the German Wikipedia using the skipgram model \cite{bojanowski2017enriching}. The pretrained vectors for FastText were obtained from \url{https://fasttext.cc/}, while the English GloVe word embedding was retrieved with the R package text2vec. The pretrained vectors for the German GloVe was obtained from \url{https://www.deepset.ai/german-word-embeddings}. We consider the 25,000 most used words according to Google books (\url{https://books.google.com/ngrams/}) in the application of word embedding spaces to lexicon expansion. 

The expansion of the set of seed words with methods based on word embedding spaces is implemented as follows: we retrieve all the words that have cosine similarity higher or equal to 0.5 to each seed word in the embedding space considered. The threshold of 0.5 on the cosine similarity has been chosen according to previous work \cite{fast2016empath,mpouli2020lexifield}. Furthermore, we consider one more elaborated method based on word embeddings: Empath \cite{fast2016empath}. This algorithm creates the extended word list by retrieving the closest words to the embedding of the cumulative vector relative to the seed words. We deploy this method via its Python package with the default size setting, which is set to an output of 100 words for the expanded word list. Since the number of words was too low for our purposes, we tried to increase this value and considered size 300, 500, 700 and 1,000. However, with these specification the Python package for Empath gave an error and did not deliver any output. Moreover, Empath is based on an old embeddings model which has not been updated since its first release. As a consequence, we decided to implement a novel version of the method and consider only that in the analyses. 

In particular, we re-implemented the Empath method using the newer FastText word embedding space trained on the English and German Wikipedia. Thus, we obtain two new versions of Empath, one for the expansion of English lexica and one for the expansion of German lexica. We call these re-implementations Empath 2.0. Note that, while Empath allows for the selection of the size of the final word list, Empath 2.0 does not have this feature because the mechanism used in Empath for discarding words was not documented in the original paper. As a consequence, the sizes of the word lists computed by the two versions of Empath differ, but we consider only Empath 2.0 as it is based on newer and more exhaustive word embedding models.

For each of the methods, we also define a baseline model based on the length of the resulting word lists. In the baseline algorithm, we perform 1,000 random samples of words from the relative networks or word embedding spaces of the same size of the expanded lexicon resulting from each method. This serves as a null model to measure what would be the performance of a random guess when expanding word lists to given sizes.

\subsection{Experiments}
  We compare the lexicon expansion methods on a lexicon expansion task in two different languages and in a text analysis exercise. In the first task, we assess the performance of the methods in expanding various word lists given a set of seed words against longer lists generated by experts. We generate the expansions of seed word lists and evaluate them against the 2015 English version \cite{pennebaker2015development} and the 2007 German version of the Linguistic Inquiry and Word Count (LIWC) \cite{wolf2008computergestutzte}. LIWC is a widely used proprietary dictionary-based method for the analysis of texts \cite{pennebaker2001linguistic}. The 2015 English version of LIWC collects 73 word lists relative to various topics, including for example words indicating future orientation or referring to the family sphere. Such word lists have been used in many influential studies, as for example \cite{kleinberg2020measuring,shing2018expert,zirikly2019clpsych}. The popularity of LIWC resulted also in its translation in various languages. In particular, we consider the German version of LIWC from 2007, which collects word lists belonging to 68 different topics. Recently, a new version of LIWC has been released \cite{boyd2022development}. Since we use LIWC only as a mean to test the performance of lexicon expansion algorithms, we do not think that the results of the paper would vastly change using the new version of LIWC.
  
  Words in both LIWC resources are given in a shortened form with wildcards (indicated with *). In a first preprocessing step, we match the words with wildcards with entries in dictionaries in order to retrieve the full-form words from the the wildcard terms. For example, the word with wildcard 'apprehens*' from the English LIWC word list for negative emotion, is matched with the following entries in the dictionary: 'apprehensible', 'apprehension', 'apprehensive', 'apprehensiveness'. We consider the lexica resulting from this matching procedure as the original LIWC word lists.
  
\begin{figure}[hbtp]
  \includegraphics[width=\linewidth]{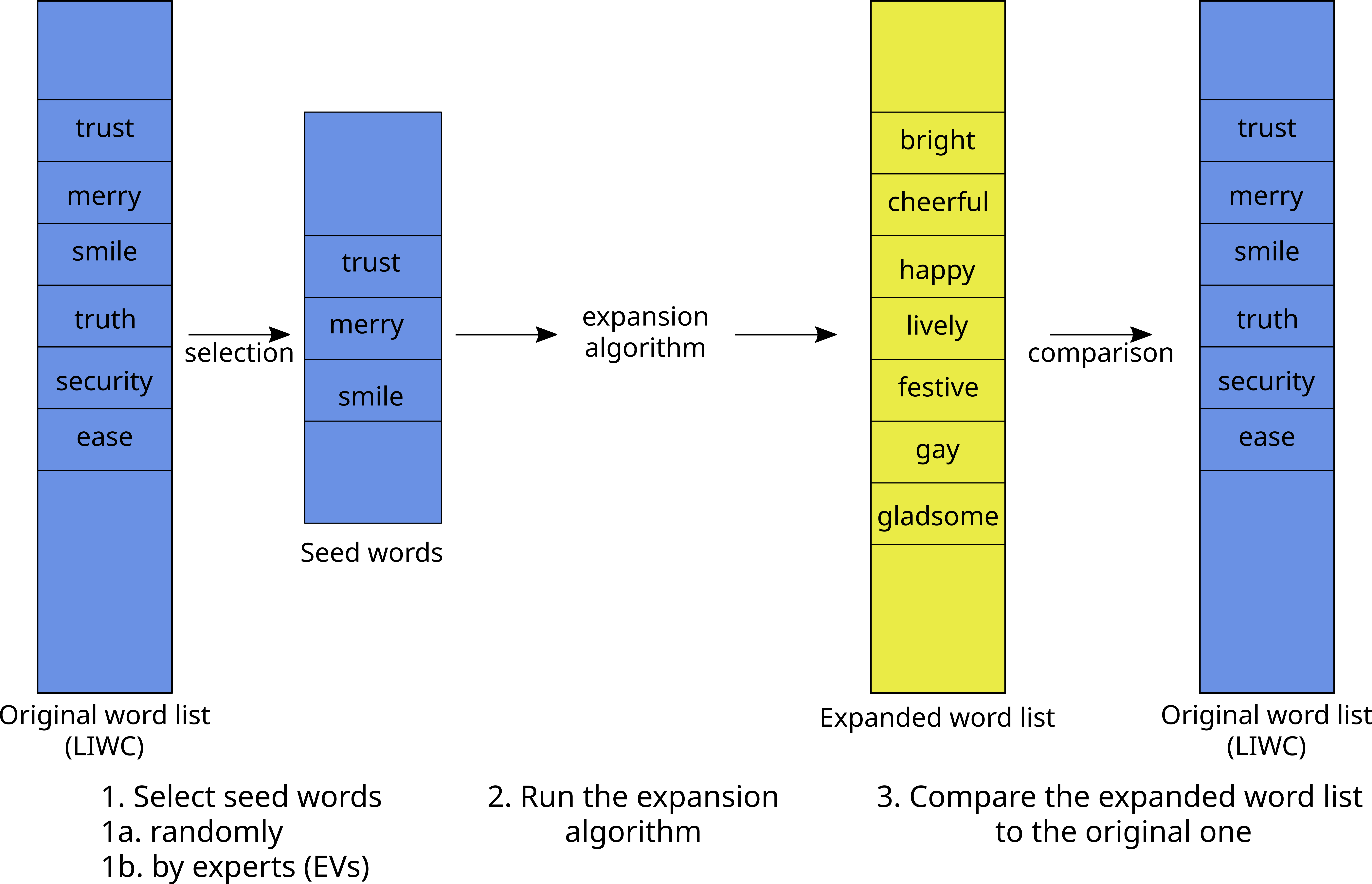}
  \caption{Representation of the word list expansion experiment. Seed words are selected from the original LIWC word list (step 1) either at random (1a) or selected by experts (1b). They are then used as input of the various expansion algorithms, which give an expanded word list as output (step 2). In order to assess the performance of each method, we compare the expanded word list with the original one (step 3) by computing the precision, recall and $F_1$.}
  \label{fig:experiment}
  \end{figure}

  After the removal of wildcards from the LIWC word lists, the experiment is performed as follows: We first select a subset of words from each word list of LIWC to use as seed words in two ways, either at random or based on expert selection of shorter lexica. We then input this seed word list in the expansion algorithms with the aim of recovering the original, complete LIWC word list. Finally, we compare the expanded and original LIWC word list and assess the performance of each method by computing the precision, recall and $F_1$ of the expanded word list. A representation of such experiment is depicted in Figure \ref{fig:experiment}.\\
  
 In particular, given $\tilde{L}$ original word list and $L = S \cup W$ expanded lexicon constructed as in equation (\ref{eq:method}), we define the true positives (TP) as the words of $\tilde{L}$ which are also present in $L$ without seed words, that is in $W$. The false positives (FP) are the words in $W$ that do not appear in $\tilde{L}$ and the false negatives (FN) are the words in $\tilde{L}$ not present in $W$. In other words:
 \begin{equation}
     TP = \tilde{L} \cap W
 \end{equation}
  \begin{equation}
     FP = W \setminus \tilde{L}
 \end{equation}
  \begin{equation}
     FN = \tilde{L} \setminus W
 \end{equation}
  We can define precision and recall as follows: \\
    \begin{equation}
      precision = \frac{|TP|}{|TP \cup FP|}
  \end{equation}
    \begin{equation}
      recall = \frac{|TP|}{|TP \cup FN|}
  \end{equation}
  $F_1$ is the harmonic mean of the previous two quantities:
    \begin{equation}
      F_1 = 2 \frac{precision \cdot recall}{precision+recall}
  \end{equation}
  
  The random selection of seed words from the LIWC lists is based on a selection of a percentage between 10\% and 90\% of the words of each word list, as depicted in Figure \ref{fig:experiment}, step 1a. We repeat the experiment 50 times for each percentage, every time selecting a new random subset of seed words. We then compute precision, recall and $F_1$ averaging on the 50 repetitions. The expert-based approach to the choice of the seed words is based on the words selected by the authors of \cite{vine2020natural} as the most representative words for the LIWC categories of negative emotion, positive emotion, anxiety/fear, anger, sadness and undifferentiated negative emotion \footnote{We decided to exclude the undifferentiated negative emotion vocabulary of the EVs from this study because it does not match any of the original thematic word list of LIWC.}. The authors of \cite{vine2020natural} call these word lists Emotional Vocabularies (EVs). Note that the EVs are freely available online, therefore we openly redistribute the expanded word lists obtained from the EVs in our work. The EVs were not created as a set of seed words from which to recover the LIWC word lists. However, we use them as they are a freely available selection of words from LIWC. The EVs are available only in English, therefore we expand them only with the English lexicon expansion methods. 

  Note that the value of precision computed in our experiments is a lower bound for the actual precision of the method: We only consider as successes the words that belong to the original LIWC word list. However, it can happen that the lexicon expansion method finds words which belong to the chosen topic but were not included in LIWC by the experts. As a consequence, in computing the precision we consider as false positives some words which might actually be true cases. The estimate of the precision is thus a lower bound, since we cannot be completely certain that LIWC presents the most extensive word lists for each topic.
  
  To complement the analysis of lower bounds, we include an analysis of additional word annotations that do not appear in the LIWC words lists and can precisely assess the true value of precision. More in detail, we generated manual annotations of the word lists resulting from the expansion of the positive and negative EVs. The first author and six other raters annotated the word lists. Five annotators are German native speakers, one speaks Italian and one Spanish as first language. They all have a near-native English proficiency. At least two annotators labeled each word in the expanded word lists and we select as relevant only the words which were accepted by both raters. In the case of FastText and Empath 2.0, the word lists resulting from the expansion procedure encompass more than 2,000 words. In such cases, we annotate a random set of 300 positive and 300 negative words per method instead of the whole word list. In all other cases, all the expanded word lists are annotated. In order to estimate the error of such statistic, we also compute the 95\% confidence interval from the bootstrapping of the annotated word lists. The inter-rater agreement relative to the annotation of the positive words scores a Cohen’s $\kappa$ of 0.59 (moderate agreement), while the task on the negative words achieves a Cohen’s $\kappa$ of 0.65 (substantial agreement). Once the word lists have been annotated, we compute a more accurate estimate of the precision of each method.
  
  Furthermore, we compare the expansion of the random selection of words from LIWC and the expansion of the EVs. This comparison has the aim of testing whether the effort of manually selecting the most representative words of one class might be an advantage to the expansion of the word list. In order to do so, we consider the expansion of a random selection of seed words of the same length of the EVs from the five emotional categories. We repeat the random selection 50 times and consider the mean performance. We compare the results in terms of precision, recall and $F_1$ with the performance of the expansion of the EVs.
  
Additionally, we analyse the interdependence of LEXpander with the other methods. In particular, we test whether the word lists created by the different methods capture different signals. In order to do so, we define a union and intersection method, which consist, respectively, in the union and intersection of the word lists resulting from the five expansion algorithms (LEXpander, WordNet, GloVe, FastText, Empath 2.0). We then analyse the performance of these methods when expanding the EVs in term of precision, recall and $F_1$. We report the results in the supplementary materials.

We continue with a second task that compares the performance of the expanded lexica in an exercise of text analysis of online communication and literary texts. In particular, we consider a simple text analysis method which consists in the computation of the frequency of words of the lexica expanded from the EVs and annotated which appear in the texts. We correlate the counts relative to each word list with the ones of the original lexica from LIWC on each single text snippet. We also compare the performance with the original EVs.

We compute the correlation of the annotated word lists obtained expanding the positive and negative EVs with the counts of LIWC on the texts of each single dataset. Since we only annotated the full-length word lists for LEXpander, WordNet and GloVe we can report the results relative to these methods. However, it is important to remember that the cleaning of the word lists was not carried out with a particular type of text in mind. This strategy would be the most advisable, but in the case of this paper we did not want to bias the results, therefore we use the same annotated word list for the four different types of texts.

For this analysis, we consider short texts of online communication from Reddit (in particular, all discussions, that is original post and answers, from the subreddits 'antiwork', 'TwoXChromosomes', 'family' and 'Home'), longer texts from the Brown corpus \cite{francis1979brown}, texts collected in the Corpus of Historical American English (COHA) \cite{davies2012expanding} and all the tweets, excluding answers, published in the UK during one single day in February 2021. The number of documents and their average length is reported in Table \ref{tab:stats_datasets}.

\begin{table}[hbtp]
    \centering
    \begin{tabular}{|c|r|r|}
    \hline
    Dataset & Num texts & Mean length \\
    \hline
    Brown corpus & 502 & 2,064 \\
    \hline
    COHA & 116,513 & 4,852\\
    \hline
    Tweets & 417,164 & 11 \\
    \hline
    Reddit & 54,499 & 1,095 \\
    \hline
    \end{tabular}
    \caption{Statistics of the datasets used for the text analysis exercise. We report the number of texts in each dataset and their average length in number of words. Stop words are included in the counts.}
    \label{tab:stats_datasets}
\end{table}

\section{Results}
In this section, we report the results of the results comparing LEXpander with other lexicon expansion methods. In the first task, we use multiple lexicon expansion methods to retrieve the original LIWC word lists from a random subset of their words. In Table \ref{tab:LIWC_all} we report the results relative to a set of seed words amounting to 30\% of the original LIWC word list. We also report the mean size of the expanded word lists and the results of the baseline methods, averaged over 1,000 repetitions. Additionally, the lengths of the expanded word lists for every experiment are featured in Tables 1, 3, 4 of the supplementary materials.
\newline
\\
\begin{minipage}{\linewidth}
    \centering
    \captionof{table}{Results of the expansion of the random choice of 30\% words from the English LIWC.}
    \begin{tabular}{|c|c|c|c|c|c|c|c|}
    \hline
         Method &\multicolumn{2}{c|}{Precision} & \multicolumn{2}{c|}{Recall} & \multicolumn{2}{c|}{$F_1$}& mean\\
         \cline{1-7}
          & mean & bl & mean & bl& mean & bl & size\\
          \hline
         LEXpander &\textbf{0.16} & 0.01 &0.14 & 0.02 & \textbf{0.13}& 0.01 &614\\
         \hline
         WordNet \cite{miller1995wordnet} &0.10 & 0.00 &0.07 &0.00 &0.07 & 0.00 &525\\
         \hline
          Empath 2.0 \cite{bojanowski2017enriching,fast2016empath} &0.08 &0.01 &0.22 & 0.03& 0.10 &0.01 & 1,293\\
         \hline
         FastText \cite{bojanowski2017enriching} &0.06 & 0.01 &\textbf{0.29} & 0.06 &0.09& 0.02 & 2,252\\
         \hline
         GloVe \cite{pennington2014glove} & 0.07 & 0.01 &0.13 & 0.03 &0.08 & 0.02 &773\\
         \hline
    \end{tabular}
\caption*{Precision, recall and $F_1$ of the expansions generated from from random 30\% seed words compared to the original lexica from the English 2015 version of LIWC. Values are means computed over 50 samples of the seed words across word lists. We also report the mean length of the expanded lexica. The results of a baseline model (bl) averaged on 1,000 repetitions of the same length are also indicated. The best performances are highlighted in boldface.}
    \label{tab:LIWC_all}
\end{minipage}
\bigskip

Table \ref{tab:LIWC_all} shows that the best precision and $F_1$ scores are reached by LEXpander, while FastText yields to the highest recall. In this setting, LEXpander does not only achieve the best precision but also the best trade-off between precision and recall. Moreover, we observe that FastText and Empath 2.0 lead to the longest word lists, with a mean length of over 1,000 words. The average length of the word lists from LIWC is 417 words, therefore FastText delivers on average more than 5 times the number of words of the original lexica. In Table 5 in the supplementary materials we include the percentage of word lists for which it was possible to compute an expansion of the word list in at least one repetition of the 50 random drawing of seed words. All the methods manage to expand all of the 73 word lists from LIWC apart from WordNet, which expands only 66 thematic categories.

\begin{figure}[hbtp]
    \centering
    \includegraphics[width=\linewidth]{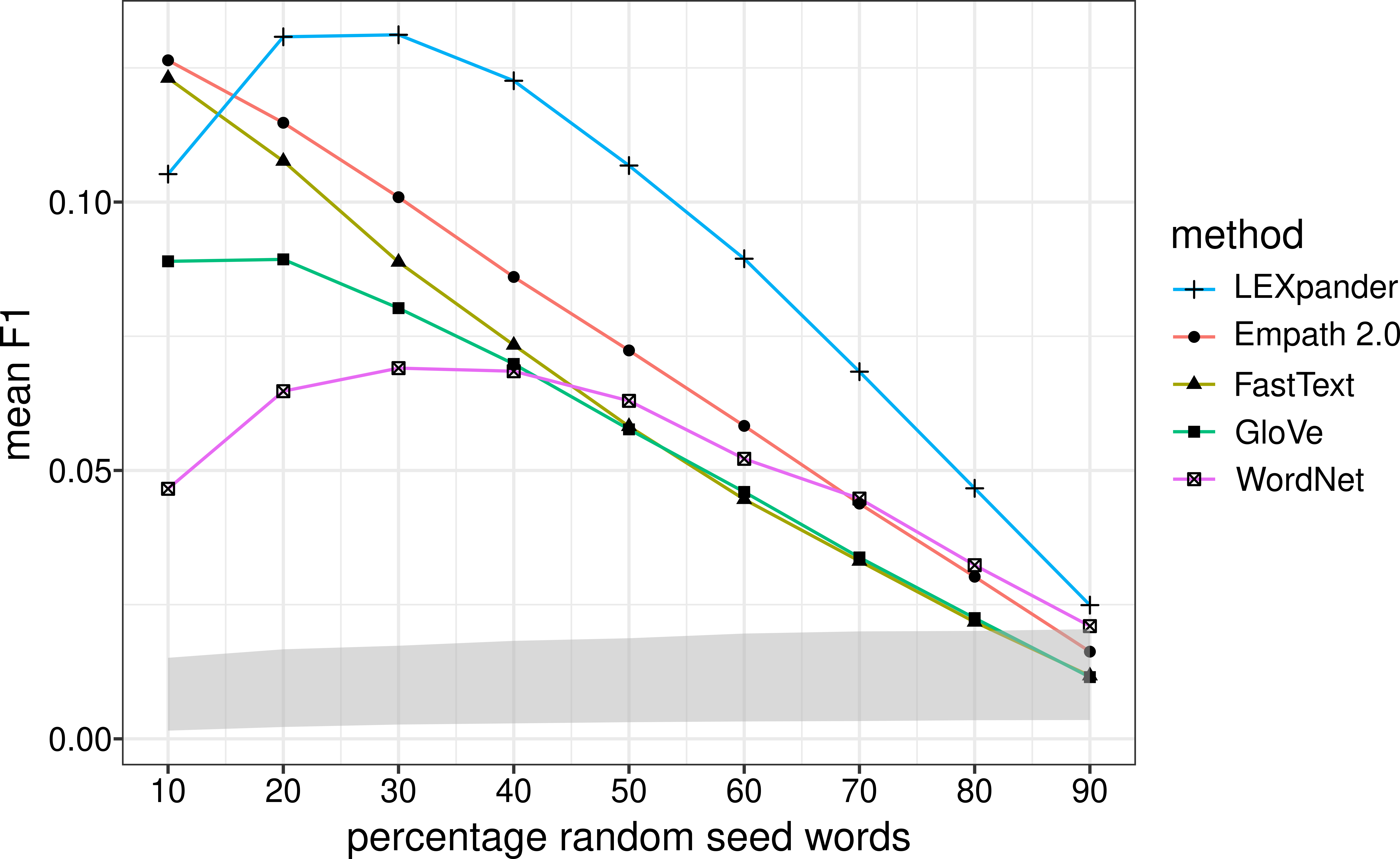}
    \caption{Mean of the $F_1$ scores of the expansion of seed words chosen at random from LIWC as a function of the percentage of seed words chosen. The mean is computed on the 73 different thematic categories. The grey area represents the baseline as the maximum and minimum mean $F_1$ of the baseline models.}
    \label{fig:F1vsth}
\end{figure}

The results of Table \ref{tab:LIWC_all} are relative to an initial set of seed words of 30\% of the LIWC word lists. In the following, we analyse the dependence of the $F_1$ value on the percentage of seed words chosen, as represented in Figure \ref{fig:F1vsth}. We also consider the values of the baseline methods as a shaded area whose borders correspond to the minimum and maximum of the mean $F_1$ scores relative to all the baseline methods.

In Figure \ref{fig:F1vsth}, we see that LEXpander achieves the best $F_1$ when considering at least 20\% words as seed words. With 10\% of seed words, Empath 2.0 and FastText have slightly better results than LEXpander and with 90\% of seed words all the methods yield to very similar performances. Therefore, LEXpander proves to be consistently the best method for the expansion of word lists for many size ranges. We also see that the fewer words are needed to recover the original LIWC word lists, the more the baseline methods outperform the actual lexicon expansion models.

These results illustrate the performance of the lexicon expansion methods when dealing with a random subset of the word lists from the English LIWC. In order to analyse the dependence of the quality of the expanded word lists on the choice of the seed words, we perform the expansion of a manual selection of words from the five emotional categories of LIWC, called EVs \cite{vine2020natural}. Similarly to the previous case, we find that LEXpander achieves the best precision and $F_1$, while FastText the best recall. The full table of results for this case is reported in Table 2 of the supplementary materials, while a concise version constitutes the left side of Table \ref{tab:comparison}. In contrast to the previous experiment, when expanding the EVs all methods achieve 100\% coverage of the word lists, that is, it was always possible to compute an expansion of the EVs.

In Table \ref{tab:comparison} we compare the performance of the methods when choosing the seed words at random from LIWC and when using a well-thought set of words to generate the final lexicon, the EVs. We consider as seed words exactly the same number of words for both cases, that is we control for the number of seed words.

\bigskip
\begin{minipage}{\linewidth}
    \centering
    \captionof{table}{Dependence of the performance of the lexicon expansion algorithms on the mode of choice of the seed words: at random or chosen by experts.}
  \begin{tabular}{|c|c|c|c||c|c|c|}
  \hline
  & \multicolumn{3}{c||}{EVs as seed words} & \multicolumn{3}{c|}{Random seed words}  \\
 \hline
 & prec &  rec &  $F_1$ & prec & rec & $F_1$ \\
 \hline
 LEXpander  & \textbf{0.16} & 0.10 & \textbf{0.12}& \textbf{0.16} (0.02) & 0.15 (0.01) &\textbf{0.15} (0.01)\\
 \hline
 WordNet \cite{miller1995wordnet}& 0.11 & 0.06 & 0.08& 0.12 (0.02) &0.08 (0.01) & 0.09 (0.01)  \\
 \hline
 Empath 2.0 \cite{bojanowski2017enriching,fast2016empath}& 0.07 & 0.29 & 0.11& 0.07 (0.00) & 0.34 (0.01) & 0.12 (0.00)  \\
 \hline
 FastText \cite{bojanowski2017enriching} & 0.06 & \textbf{0.34} & 0.10& 0.07 (0.00) & \textbf{0.40} (0.01) & 0.11 (0.01) \\
 \hline
 GloVe \cite{pennington2014glove}& 0.07 & 0.03 &0.04& 0.06 (0.01) & 0.04 (0.01) & 0.04 (0.01) \\
 \hline 
 \end{tabular}
 \caption*{Mean of precision, recall and $F_1$ on the five emotional categories either choosing the seed words at random from the relative LIWC dictionaries or manually by experts (EVs). The standard deviation on 50 repetitions of the random choice of seed words is reported in between parentheses. We control for the length of the seed words.}
 \label{tab:comparison}
 \end{minipage}
 \bigskip

In Table \ref{tab:comparison}, we see that precision, recall and $F_1$ values are always higher or equal when taking a random sample of words from LIWC than when expanding a selection of words made by experts. This might be due to the fact that we repeat the random choice of seed words from LIWC 50 times, averaging the estimates for precision, recall and $F_1$. Therefore, even if one random subset of seed words is not fully representative for the theme of the word list to reconstruct, it might be balanced by the other random choices. However, the standard deviation relative to the means of precision, recall and $F_1$ on the 50 repetitions show that the variability in the results is minimal. One alternative explanation for the observation is that the EVs were not intended to be used to reconstruct the original LIWC. Rather, the aim of their creation was to quantify the vocabulary width of people with respect to emotions. Therefore, they might collect very frequent words, while the original LIWC word lists might have a better distribution with respect to word frequency. Note that, even when the seed words were chosen at random, they were anyways selected from a thematic set of words, that is words that convey a specific meaning. Thus, this comparison does not prove that the seed words do not have to be relevant to the desired topic, but that they do not have to be the most fitting and representative ones.

We also consider the union and intersection of the expanded word lists in order to determine whether a combination of the lexica leads to better results. We find that the union model scores a precision value of 0.15 and a recall value of 0.04, thus leading to a $F_1$ score of 0.06. The intersection yields to a very high precision (0.75), but the recall in 0, thus the $F_1$ score relative to the method is 0. Therefore, the only case in which one of the combinations outperform the best scoring method consists in the precision results of the intersection model. However, this method cannot compete with the single ones with respect to recall and $F_1$. Thus, we can conclude that the intersection and union of the word lists does not yield to better results.

In Table 2 of the supplementary materials we compute the precision, recall and $F_1$ score of the EVs over the original LIWC resource. Since the precision is lower than 1 (in particular, 0.86), we observe that the creators of the EVs included some words that do not appear in the original vocabulary. Thus, we  can conclude that it is possible to add relevant words to LIWC, that is LIWC does not cover all the words relative to a topic. As a consequence, the precision we computed when comparing the expanded and original LIWC word lists is a lower bound for its real value. In the following precision study we analyse this difference by collecting manual annotations of the word lists generated expanding the EVs. We report the lower bound for precision (indicated with *) and the estimate of its true value with respect to the annotations in Table \ref{tab:precision_study}.\\

\bigskip
\begin{minipage}{\linewidth}
    \centering
    \captionof{table}{Precision study of the lexicon expansion methods when using the EVs as seed words.}
    \begin{tabular}{|c|c|c|c|c|}
    \hline
     & \multicolumn{2}{c|}{Precision*} & \multicolumn{2}{c|}{Precision}\\
    \hline
         & Negative & Positive & Negative & Positive\\
       \hline
         LEXpander & \textbf{0.21} & \textbf{0.20} & \textbf{0.64} [0.61,0.67] &\textbf{0.43} [0.40,0.47]\\
         \hline
          WordNet \cite{miller1995wordnet}& 0.15&0.11 & 0.63 [0.60,0.67]& 0.41 [0.40,0.47]\\
         \hline
         Empath 2.0 \cite{bojanowski2017enriching, fast2016empath}& 0.13 & 0.10 &0.47 [0.41,0.52]&0.35 [0.30,0.40]\\
         \hline
        FastText \cite{bojanowski2017enriching}& 0.10&0.09 &  0.41 [0.36,0.47]&0.28  [0.23,0.33]\\
         \hline
         GloVe \cite{pennington2014glove} & 0.11&0.10 & 0.25 [0.21,0.30]& 0.18 [0.15,0.21]\\
         \hline
    \end{tabular}
    \caption*{Comparison of the lower bound for precision (indicated with *) with the precision value adjusted according to the annotations of raters. We include 95\% confidence intervals for the estimate for true precision. In bold the best results for each computation are reported.}
    \label{tab:precision_study}
\end{minipage}
\bigskip

From Table \ref{tab:precision_study}, we see that LEXpander achieves the highest precision value for both positive and negative word lists, as also highlighted in previous results (see Table \ref{tab:LIWC_all}). However, the estimated real precision of LEXpander is not statistically different from the one of WordNet, and the two methods perform significantly better than the other models. Therefore, methods based on word networks outperform the ones constructed on word embeddings with regard to the precision of the expansion of lexica. However, the recall score of WordNet is markedly lower than the one of LEXpander, therefore we can assume that the latter continues to score the best $F_1$ value. 

The adjusted precision reported in Table \ref{tab:precision_study} is always at least 1.8 times higher than the lower bound for precision, thus corroborating the idea that the precision we could compute given the word lists from LIWC was only a lower bound. Moreover, the correlation between the lower bound and the adjusted precision values is 0.71 ($p=0.02$). It is interesting to observe that a low value for precision does not always imply a low value in the adjusted precision: for example, in the case of the positive emotion category, methods with an estimated value for precision smaller than 0.12 yield to an estimated true value between 0.18 (GloVe) and 0.41 (WordNet). As a consequence of the new estimates for precision, we can conclude that also the values estimated for $F_1$ in the previous tasks are lower bounds and close annotation tasks like this one reveal that the true performance is higher.

We then test the performance of LEXpander and the other lexicon expansion algorithms in the German linguistic setting. In order to do so, we expand a random selection of seed words of the German version of LIWC from 2007. In Table \ref{tab:LIWCdeu_all} we report the results relative to a 30\% random selection of words.\\

\begin{minipage}{\linewidth}
    \centering
    \captionof{table}{Results of the lexicon expansion task with a random selection of 30\% words from the lexica of the German LIWC.}
    \centering
    \begin{tabular}{|c|c|c|c|c|c|c|c|}
    \hline
         Method & \multicolumn{2}{c|}{Precision} & \multicolumn{2}{c|}{Recall} & \multicolumn{2}{c|}{$F_1$}&mean\\
         \cline{1-7}
          & mean & bl & mean & bl & mean & bl & size \\
         \hline
         LEXpander & \textbf{0.07} & 0.02 & \textbf{0.20} & 0.05 & \textbf{0.09} & 0.02 &1,714\\
         \hline
         OdeNet \cite{siegel2021odenet} & 0.03 &0.00 & 0.00 & 0.00 & 0.00 & 0.00 & 170\\
         \hline
         Empath 2.0 \cite{bojanowski2017enriching,fast2016empath}& 0.03 & 0.01 &0.14 & 0.02 & 0.04 & 0.01 &1,905\\
         \hline
         FastText \cite{bojanowski2017enriching}& 0.03 & 0.01 &0.16&0.03 & 0.04&0.01 & 2,350\\
         \hline
         GloVe \cite{pennington2014glove}&0.05 &0.01 & 0.13 &0.02& 0.05 & 0.01& 722\\
         \hline
    \end{tabular}
    \caption*{Precision, recall and $F_1$ of the word list retrieved with 30\% of seed word chosen at random versus the original word lists from the German LIWC. We report the performance of the relative baseline method (bl) and the mean size of the expanded word lists. In bold are highlighted the best performances.}
    \label{tab:LIWCdeu_all}
\end{minipage}
\bigskip

In Table \ref{tab:LIWCdeu_all} we see that LEXpander yields to the best values for precision, recall and $F_1$ in the German setting, thus confirming the trend already observed with English in Table \ref{tab:LIWC_all}: LEXpander features the best precision and the best trade-off between recall and precision overall. Moreover, because of the low results of the other methods in this linguistic setting, LEXpander yields also to the best recall. We observe that in general the mean size of the final word lists (see Table 4 of the supplementary materials) is larger than the one obtained in the English setting. This is probably a result of the fact that German has more word inflections than English. Also in this case we find that FastText and Empath 2.0 deliver the largest word lists but the discrepancy in length with the word lists expanded with the other methods is less dramatic than in the previous cases.

\begin{figure}[hbtp]
    \centering
    \includegraphics[width=\linewidth]{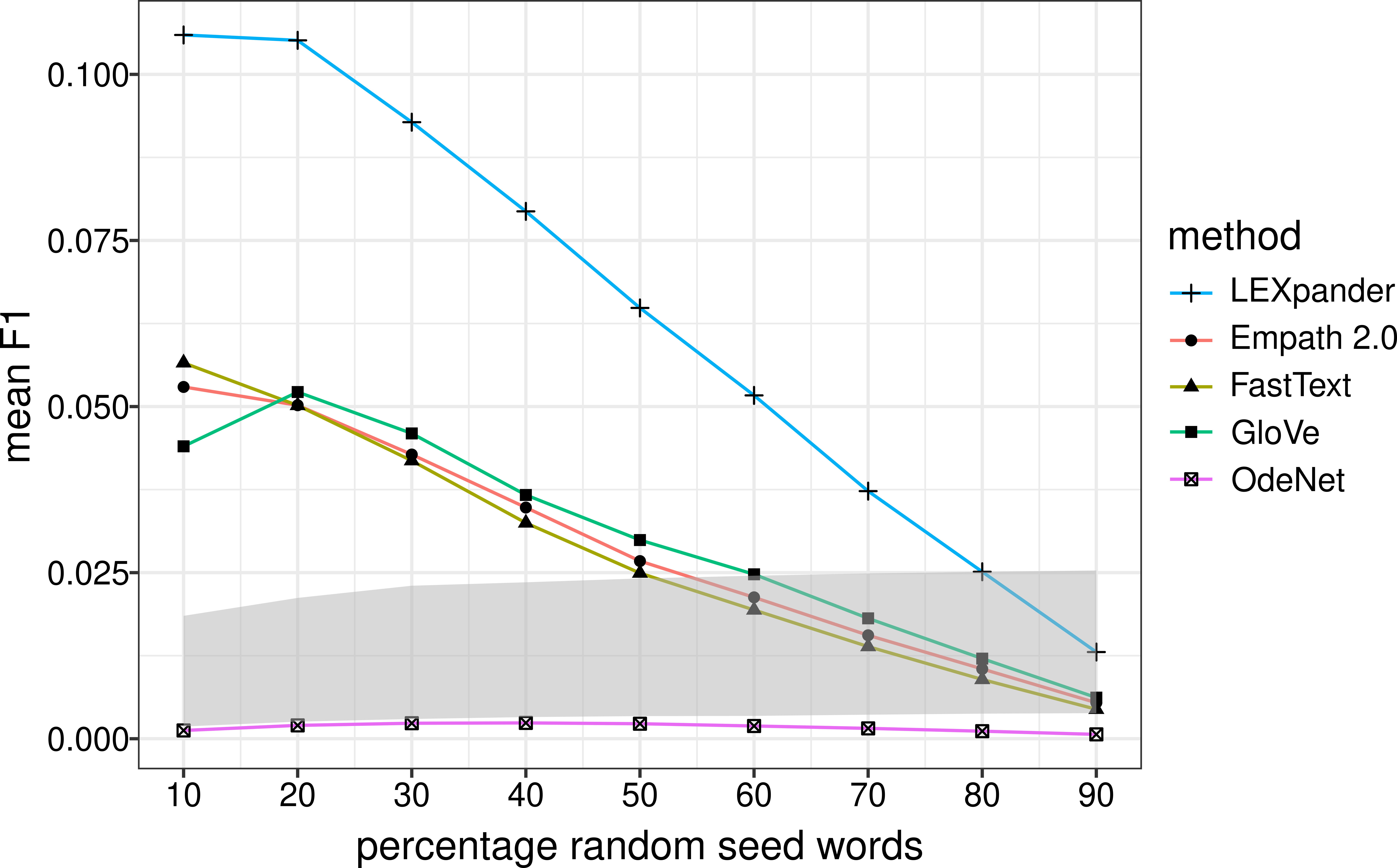}
    \caption{Mean of the $F_1$ scores of the expansion of the German LIWC as a function of the percentage of words chosen at random from the original lexicon. The grey area represents the maximum and minimum of the baseline models.}
    \label{fig:F1vsth_de}
\end{figure}

Also in the German case, we consider the value of $F_1$ as a variable of the number of seed words, as represented in Figure \ref{fig:F1vsth_de}.

Figure \ref{fig:F1vsth_de} shows that LEXpander reaches the best $F_1$ for any size of seed words chosen. Moreover, the $F_1$ results and the relative difference in performance of the methods decreases the more seed words are considered, while the results of the baseline methods increase. We observed the same pattern in the case of the English LIWC (see Figure \ref{fig:F1vsth}). This happens because very few words have to be added to the expanded word list in order to recover the original one. We also observe that OdeNet delivers worse results than the baseline model, i.e., a random model achieves better $F_1$ than OdeNet. This is probably due to the limited size of the OdeNet network as hinted by the number of word lists the method manages to expand: 27 out of the 68 word lists in the German LIWC. All the other method expand more than 60 word lists and LEXpander achieves the highest number of expanded word lists: 66 out of 68. These results are reported in Table 5 of the supplementary materials.

\begin{figure}[hbtp]
  \includegraphics[width=\linewidth]{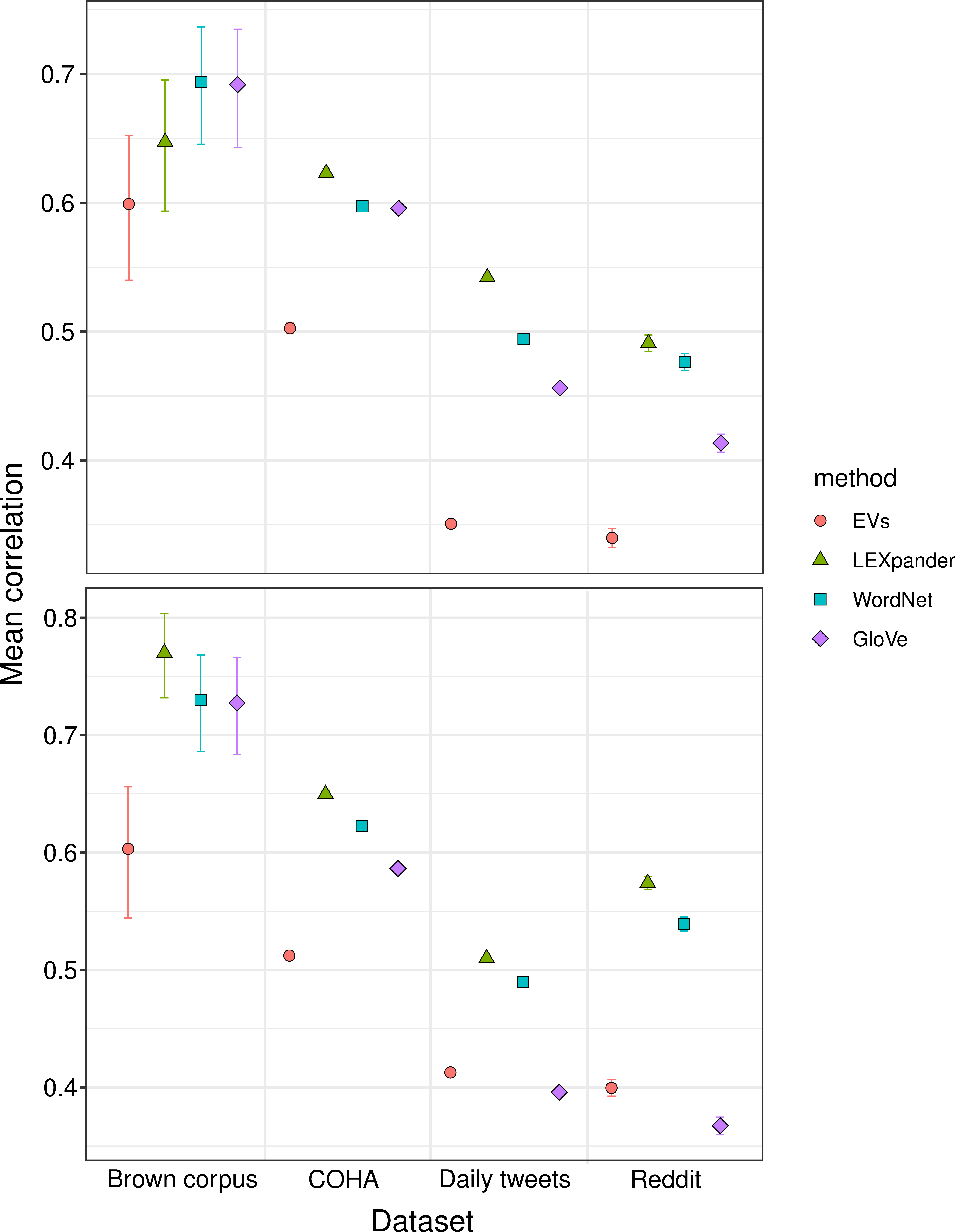}
  \caption{Correlation of the frequency of words in the positive (top) and negative (bottom) expanded word lists from the EVs and the ones from LIWC in texts from different datasets. The performances of the EVs is considered as a baseline. The bars indicate the 95\% confidence intervals. In some cases, error bars are narrower than point size.}
  \label{fig:corr_EV}
  \end{figure}

As a last test, we perform a text analysis validation task on long and short texts from online and offline communication. We compare the frequencies computed using the expanded word lists with the ones of the EVs and the original LIWC word lists on each dataset. We consider the annotated word lists from one of the previous tests, which were expanded from the positive and negative EVs using LEXpander, GloVe and WordNet and report the correlations in Figure \ref{fig:corr_EV}.

In the text analysis exercise we see that LEXpander always achieves best or is tied with the best correlation on all the datasets. In particular, LEXpander, WordNet and GloVe yield to statistically indistinguishable results from the ones of the EVs on the Brown corpus for the positive lexicon. With respect to negative sentiment, the three models are indistinguishable but significantly outperform the baseline given by EVs. On the other datasets, namely COHA, Reddit and the tweets published on one day, LEXpander achieves the best correlation with the positive and negative word lists from LIWC, outperforming also the EVs.

\section{Discussion}
In this paper, we present a new lexicon expansion algorithm, LEXpander, and compare its performance in a benchmark including different automatic lexicon expansion algorithms. We show that LEXpander achieves the best precision and $F_1$ in the lexicon expansion tasks in two linguistic settings, as well as best or tie with the best in a text analysis exercise. LEXpander is an open source method available as a web tool (\url{https://annadinatale.shinyapps.io/lexpander_app/}). Moreover, the word lists expanded from the EVs are shared on the GitHub page (\url{https://github.com/AnnaDiNatale/LEXpander}), as well as the code realised along with the present paper. LEXpander is a lexicon expansion algorithm based on a linguistic principle, colexification, and the present publication shows the usefulness of bridging linguistic theory and NLP applications. Incorporating linguistic theory can provide novel, interpretable models, which give insights into phenomena rather than fitting statistical features of texts with black box algorithms.

Our work also shows that some linguistic ideas can solve the problem of English-centric research. Indeed, when deploying methods that are independent from language, as colexification networks, we showed that the resulting method has good performances in two different linguistic settings. In particular, the quality of the results of LEXpander in the German setting is more pronounced than the one when considering English. This is because all the other word lists expansion methods considered were developed and validated taking into account only the English language. By making use of a language-independent concept, colexification, we show the potential applications enabled by this property of the method.

In addition to this, we find that lexicon expansion methods based on networks outperform the ones based on word embeddings in terms of precision and $F_1$. We also proved that the union of the expanded word lists does not yield to better results. These evidences are in contrast with the ones of \cite{bozarth2022keyword}, which are relative to the amount of relevant tweets key word expansion algorithms can retrieve. Indeed, the focus of the two papers is different as our analysis addresses a different research question and does not deal with the problem of text mining on Twitter.

In the word list expansion experiments we performed, we saw a substantial difference in the length of the expanded word lists, especially with respect to FastText and Empath 2.0. The length of the expanded word list increased the recall of the method but reduced its precision, thus leading to low $F_1$ scores. This evidence suggests that the cosine similarity parameter threshold of methods based on word embeddings is not negligible. To our knowledge, an analysis of the dependence of word embedding methods on the cosine similarity threshold used to select words is missing. However, this analysis could result in methods with enhanced performance. Since a higher cosine similarity threshold translates in a smaller word list, under this configuration the model might deliver better results.

While LEXpander and other lexicon expansion methods offer an easy way to improve word lists, we do not recommend using them without some degree of manual inspection and filtering. Such selection should be performed taking into account the application and the type of language of the texts considered. For example, in the text analysis exercise we showed that the positive and negative word lists obtained with LEXpander have the highest correlation with LIWC on all the dataset considered. However, the annotation of the expanded word lists had been performed without the aim of performing such an application. Therefore, we think that the performance of all the methods would have been better if the cleaning process would have been performed with a specific dataset in mind.

Moreover, there are some cases where researchers might want to use other methods than LEXpander, especially when focusing on niche domains and contexts. In particular, LEXpander does not allow to explore novel ways of usage of language in a specific linguistic environment, as word embeddings do when trained on novel corpora. Therefore, when exploring the language used in a medium to find patterns never analysed before, as in \cite{balsamo2021patterns}, word embeddings seem to offer a better alternative. On the contrary, LEXpander is the method to use in the pre-processing of data, as for example when selecting text instances relative to a topic or when brainstorming for the creation of word lists related to general topics. Moreover, the multilingual feature of LEXpander makes it the preferable choice for these tasks when considering languages with lower resources compared to English.

LEXpander is a resource that can be used both when brainstorming and compiling lexica and when doing text analysis after an adequate cleaning. Future work with LEXpander might include the analysis of psychological phenomena with the help of this resource. For example, the dictionary from the Moral Foundation Theory \cite{graham2009liberals,graham2013moral} might be expanded for better capturing signals in texts. Another application might rely on the creation of novel word lists, as for example the ones intended to test new concepts from psychology or sociology, as the ideas of loose and tight cultures in \cite{jackson2019loosening}.

 To sum up, we find that LEXpander combines a high coverage of the thematic categories and the best trade-off between precision and recall in the task of expanding a word list both in English and German. The absolute values of the performance might be deceivingly low: the best method, LEXpander, achieves a $F_1$ score of 0.12 when expanding the EVs. However, this value represents only a lower bound for the actual $F_1$ score, as proven by the precision study. Indeed, in such test we prove that the precision value we compute is a lower bound and that in the case of LEXpander the real precision value is at least 2 times higher than its lower bound.

\section{Conclusion}
In this paper, we introduce a novel lexicon expansion algorithm, LEXpander. LEXpander implements a method based on a colexification network, that is a multilingual semantic network. We test the performance of LEXpander on various lexicon expansion tasks, comparing it to other widely used lexicon expansion algorithms, including methods based on the GloVe and FastText word embeddings, and algorithms deploying semantic networks. We find that LEXpander is the best option when focusing on precision or $F_1$ in English and the best method overall in the German settings. The German experiment shows the performance of the method in a non-English setting, but the tool can be applied other languages including Spanish, Swahili, Croatian and Swedish.

Even if LIWC might seem the best method for some tasks, such tool is not open source, therefore an alternative method might be more widely used. A freely available tool is Empath \cite{fast2016empath}, often deployed in research thanks to its ease of use. However, such method is now outdated and delivers very short expanded word lists. LEXpander is a free, open-source multilingual tool  that can be found in a GitHub repository and can be used through an interactive page. Moreover, we share the word lists obtained from the expansion of the EVs on GitHub (\url{https://github.com/AnnaDiNatale/LEXpander/tree/main/expanded_wordlists/Annotated}), which provide a resource comparable to LIWC word lists without using any LIWC dictionary data. These resources are freely available for download, with the aim of supporting future research on text analysis methods and their applications.

\section*{Acknowledgments}
We thank Emma Fraxanet, Alina Herderich, Jana Lasser, Brigitte Marti, Hannah Metzler, and Max Pellert for the annotations of word lists generated by LEXpander. We are also grateful to Aleszu Bajak for the help with creating the LEXpander web tool.
\section*{Declarations}
\subsection*{Funding}
This study was founded by the Vienna Science and Technology Fund through the project “Emotional Well-Being in the Digital Society” (Grant No. VRG16-005).
\subsection*{Competing interests}
The authors have no competing interests to declare that are relevant to the content of this article.
\subsection{Data availability}
Datasets and codes for reproducing our results  are available in the GitHub repository, \url{https://github.com/AnnaDiNatale/LEXpander}, with the exception of LIWC word lists, which can be purchased by any researcher from \url{https://www.liwc.app/}.

\section{Supplementary materials}

In the first task we expand a set of seed words into a word list. Expanding a random sample of 30\% of the words in each category of the English LIWC, we find word lists with different lengths. Table \ref{tab:lengths} reports the lengths of the final word lists for the 5 emotional categories and the mean length on all the 73 categories in comparison to the length of the original lexica from LIWC.

\begin{table}[hbtp]
    \centering
    \begin{tabular}{|c|c|c|c|c|c|c|c|}
    \hline
         & LIWC & LEXpander & WordNet &Empath 2.0  & FastText & GloVe \\
         \hline
         Negemo & 1,410  & 1,626 &1,222& 3,227&5,916 &1,873 \\
         \hline
         Posemo & 1,052 & 1,966 & 1,839  &4,019& 6,977&1,613 \\
         \hline
         Anx & 263 & 428&331& 3,170& 3,681&311  \\
         \hline
         Anger & 455 & 656  &668 & 3,020& 4,201&516 \\
         \hline
         Sad & 258 & 464 &327 & 2,862& 3,333&440  \\
         \hline
         \hline
        mean & 417 & 614 &525 & 1,293&2,252 &773  \\
         \hline
    \end{tabular}
    \caption{Mean length of the expanded word lists with 30\% random words from LIWC as seed words. We report the mean lengths of the emotional categories on 50 repetitions and the mean over all the categories.}
    \label{tab:lengths}
\end{table}

Table \ref{tab:lengths} shows that the word lists obtained with FastText and Empath 2.0 are the longest, consisting on average of nearly 3 times more words than the original lexicon.

\bigskip
\begin{minipage}{\linewidth}
    \centering
    \captionof{table}{Results of the expansion of the EVs.}
  \begin{tabular}{|c|c|c|c|c|c|c|c|}
  \hline
  Method &\multicolumn{2}{c|}{Precision} & \multicolumn{2}{c|}{Recall} & \multicolumn{2}{c|}{$F_1$}& mean\\
  \cline{1-7}
          & mean & bl & mean & bl & mean & bl & size\\
 \hline
 Original EVs & 0.86 & - & 0.19 & - & 0.30&- & 132\\
 \hline
 \hline
 LEXpander & \textbf{0.16} & 0.02 & 0.10 & 0.01 & \textbf{0.12} & 0.02 &570\\
 \hline
 WordNet & 0.11 & 0.00 &0.06 & 0.00 &0.08 & 0.00 &492\\
 \hline
 Empath 2.0 & 0.07 & 0.02 & 0.29 & 0.07 & 0.11 & 0.03 &2,702\\
 \hline
 FastText & 0.06 &0.02 & \textbf{0.34}&0.10 & 0.10 & 0.03 &3,684\\
 \hline
 GloVe & 0.07& 0.01& 0.03&0.01 &0.04 &0.01 & 419\\
 \hline 
 \end{tabular}
 \caption*{Mean of precision, recall and $F_1$ of the expansion of the EVs in comparison to the relative word lists of LIWC. The results of the baseline models (bl) are also reported. The mean of the performances is computed on the five emotional word lists. The best results are indicated with boldface. In this case, we also report the comparison of the original EV wordlists with the lexica from LIWC (first row).}
 \label{tab:expansion_EV}
 \end{minipage}
 \bigskip
 
In the comparison between the EVs and LIWC we observe that the precision is lower than 1, which would have been expected if the EVs would have been created choosing only words from LIWC. Therefore, when creating the EVs, researchers added some words which were not present in the original LIWC word lists. This hints to the fact that LIWC is not the most extensive source, at least for emotional word lists.

In Table \ref{tab:length_EV} we report the length of the word lists obtained expanding the EVs. Also in this case, FastText and Empath 2.0 yield to the largest lexica.\\
\begin{table}[hbtp]
    \centering
    \begin{tabular}{|c|c|c|c|c|c|c|c|}
    \hline
        length & EVs & LEXpander & WordNet &Empath2.0& FastText &GloVe\\
         \hline
         Negemo & 276 & 1,068 & 979&  3,325 & 5,288 & 672 \\
         \hline
         Posemo & 172 & 815 &685  &  2,723 & 3,835& 878\\
         \hline
         AnxFear & 62 & 241 &194 & 2,579 & 3,312& 186\\
         \hline
         Anger & 55 & 285  &279  & 2,417& 2,960& 128\\
         \hline
         Sad & 95 & 443  &325  & 2,468& 3,023& 229\\
         \hline
         \hline
        mean & 132 & 570 &492 & 2,702 & 3,683 & 419 \\
         \hline
    \end{tabular}
    \caption{Length of the expanded word lists using the EVs as seed words. The mean is computed on the 5 emotional categories.}
    \label{tab:length_EV}
\end{table}

We then expand a random selection of seed words from the German LIWC in order to test whether the methods yield to comparable results in a different linguistic setting. The mean length of the word lists obtained with 30\% of seed words is reported in Table \ref{tab:lengths_deu}.

Also in this case, both FastText and Empath 2.0 output large word lists, but the discrepancy from the other methods is less dramatic than in the case of English. On the opposite, OdeNet yields to the shortest word lists and does not manage to expand the word list for Anger.

\begin{table}[hbtp]
    \centering
    \begin{tabular}{|c|c|c|c|c|c|c|}
    \hline
         & LIWC deu& LEXpander& OdeNet &Empath 2.0  & FastText &GloVe\\
         \hline
         Negemo & 2,130 & 4,501&676 & 4,900 & 8,815 & 1,475\\
         \hline
         Posemo & 1,576 & 5,385& 503 & 4,656 & 7,394 & 1,791\\
         \hline
         Anx & 276 & 1,057& 94 &3,138& 2,948  & 204 \\
         \hline
         Anger & 570 & 1,559& 169 & 3,123 & 3,620  & 361\\
         \hline
         Sad & 462 & 1,404& 147 & 3,728& 3,619   & 464 \\
         \hline
         \hline
        mean & 528 & 1,714& 170 & 1,905& 2,744   & 722 \\
         \hline
    \end{tabular}
    \caption{Length of the expanded word lists with 30\% random words from the German LIWC as seed words. We report the mean length of the emotional categories over 50 repetitions of the random choice and the mean of the lengths over all the categories.}
    \label{tab:lengths_deu}
\end{table}

Some of the methods might not be able to extend some word lists. Indeed, in the case of methods based on networks, it might not be possible to match the seed words on the network or those words might be part of an unconnected component. In the case of word embeddings, the seed words might be isolated from other words. In Table \ref{tab:coverage} we report the percentage of wordlists which could be expanded with each method. In the case of the random selection of seed words (LIWC en and LIWC deu) we consider the percentage of wordlists which could be expanded in at least one of the 50 random sampling of the seed words.\\
\begin{table}[hbtp]
    \centering
    \begin{tabular}{|c|c|c|c|}
    \hline
        % & \multicolumn{3}{c|}{Percentage word lists expanded} \\
        % \hline
         & EVs & LIWC en & LIWC deu \\
         \hline
         LEXpander & 100\% & 100\% & 97\% \\
         \hline
         WordNet & 100\% & 92\% & - \\
         \hline
         OdeNet & - & - & 40\% \\ 
         \hline
     Empath 2.0 & 100\% & 100\% & 94\%\\
         \hline
         FastText & 100\% & 100\% & 94\% \\
         \hline
         GloVe & 100\% & 100\% & 88\% \\
         \hline
    \end{tabular}
    \caption{Percentage of the word lists for which it was possible to compute an expanded version. In the case of LIWC in English and German, the case with 30\% random words is considered. Some methods could not be applied to word lists in one language. We use the symbol - to indicate such cases.}
    \label{tab:coverage}
\end{table}

From Table \ref{tab:coverage} we see that the EVs could always be expanded by all the methods. On the opposite, the selections of 30\% random words from the LIWC lexica in English and German show a different score. In particular, WordNet and OdeNet have the lowest score respectively on LIWC in English and in German. Nearly all the methods could recover 100\% of the wordlists from the English LIWC, while none of them has the same performance on the German lexica, where LEXpander reaches the highest percentage, 97\%.

 \bibliographystyle{unsrtnat} 
\bibliography{bibliography} %Prints bibliography
\end{document}